\documentclass[tikz, 10pt]{amsart}
\usepackage[usenames,dvipsnames]{xcolor}
\usepackage{amsmath,amsthm,amssymb,color,comment,csquotes,enumerate,fancyhdr,filecontents,graphicx,verbatim}
\usepackage[round]{natbib}
\usepackage{pgfplots}
\usepackage{caption}

\usepackage{hyperref}
\hypersetup{colorlinks=true,linkcolor=MidnightBlue,citecolor=MidnightBlue,bookmarks=false,backref=page}

\usepackage{pgfplots}

\usepackage{caption}

\usepackage{pgfplots}
\pgfplotsset{compat=1.18}

\usepackage{cleveref}
\crefname{section}{§\hspace{-0.1cm}}{§§}
\Crefname{section}{§}{§§}

\makeatletter
\newcommand{\customlabel}[2]{%
   \protected@write \@auxout {}{\string \newlabel {#1}{{#2}{\thepage}{#2}{#1}{}} }%
   \hypertarget{#1}{#2}
}
\makeatother

\DeclareMathOperator*{\argmin}{arg\,min}

\usepackage[colorinlistoftodos,prependcaption]{todonotes}

\title{Benign Interpolation and Occam's Razor}
\author[Sterkenburg]{Tom F.\ Sterkenburg}
\address{Munich Center for Mathematical Philosophy (MCMP), LMU Munich \newline \indent Munich Center for Machine Learning (MCML)}
\email{tom.sterkenburg@lmu.de}
\author[Herrmann]{Daniel A.\ Herrmann}
\address{Department of Philosophy, University of North Carolina, Chapel Hill}
\email{danher@unc.edu}
\author[Romeijn]{Jan-Willem Romeijn}
\address{Faculty of Philosophy, University of Groningen}
\email{j.w.romeijn@rug.nl}
\date{\today. This is a preliminary version. We welcome feedback.}

\begin{document}

\maketitle

\begin{abstract}
    \noindent Contemporary deep learning methods generalize well even when they fit their training data perfectly, a phenomenon known as benign interpolation. This phenomenon cannot be accounted for by classical statistical learning theory and has prompted a range of attempted new explanations in the statistics and machine learning literature. A common feature of these new proposals is an appeal to a simplicity preference among interpolating models, often presented as a form of Occam's razor. We clarify this debate for a philosophical audience and argue that this new appeal to simplicity creates an explanatory gap. The classical theory offers theorems which connect the simplicity of model classes to good generalization, thus underwriting methodological simplicity norms. The new accounts instead appeal to properties of individual models, which they interpret as a kind of simplicity. Lacking a provable connection to generalization, it is the name "simplicity" that does the work a theorem used to do, making a substantive and unargued assumption look like the application of a familiar methodological principle.   
\end{abstract}

\small

\begin{quote}
\textit{These images were false for another reason also; namely, that they were necessarily much simplified\ldots{} Perhaps, indeed, the enforced simplicity of these images was one of the reasons for the hold that they had over me.}

\centerline{\textit{. . .}}

\textit{But when a belief vanishes, there survives it---more and more ardently, so as to cloak the absence of the power, now lost to us, of imparting reality to new phenomena---an idolatrous attachment to the old things which our belief in them did once animate.} \\
---{\footnotesize\citeauthor{Pro22swann} (\citeyear{Pro22swann}) \emph{Swann's Way}, ``Place-Names: The Name,'' trans.\ C. K. Scott Moncrieff}
\end{quote}

\normalsize

\section{Introduction}
 The predictive success of deep learning is the topic of heated debate among computer scientists and mathematical statisticians. The standard treatment in terms of statistical learning theory (SLT) does not suffice to explain it. SLT bounds predictive error on the basis of restrictions on the class of  models a learning algorithm can return, but modern deep learning networks work with model classes so large that these bounds become vacuous. Deep learning methods achieve perfect fit on training data while still generalizing well, a phenomenon known as \emph{benign interpolation}. This runs against the classical statistical intuition that perfect fit to training data signals overfitting. Thus the \emph{generalization puzzle} \citep{BGKP22inc}: what explains the predictive success of these methods?\footnote{Several authors have before noted the philosophical interest of this debate \citep{Ste18reas,SteGru21syn,Rae22pos,BucRai25bjps,GroGenSul24pc,BatWoo25arx}.}
 
A prominent response in the  literature appeals to a built-in simplicity preference: among the many models that fit the training data perfectly, the learning algorithm is said to prefer those that are \textit{simplest}. And simplicity promotes successful generalization. This looks like a straightforward appeal to a methodological principle that was already central to the classical theory: the principle to prefer simplicity, and trade it off against fit, often referred to as Occam's razor. 

In this paper we argue that the simplicity at work in these new accounts is categorically different from the simplicity preference that SLT underwrites. The classical theory gives us a provable connection between the simplicity of a model class and generalization. The new accounts instead appeal to properties of individual models, like low norm or descriptional complexity, which they \textit{interpret} as a kind of simplicity. But here no theorem connects those properties to generalization. With no such theorem to lean on, it is the name "simplicity" that does the work the theorem used to do, making an unargued and substantive assumption look like the application of a familiar methodological principle. This leaves a gap in the explanation of benign interpolation. 


The plan is as follows. Section \ref{sec:classical} reviews SLT and the simplicity norms it underwrites. Section \ref{sec:Generalization} lays out the puzzle of benign interpolation. Section \ref{sec:implicit} discusses the basic strategy of accounts that invoke simplicity of individual models to resolve it. Section \ref{sec:JustNewOccam} analyzes the explanatory gap that this basic strategy leaves, and articulates some possible general directions to try and fill it.

\section{The classical theory and Occam's razor}
\label{sec:classical}

Here we discuss the standard theoretical framework for machine learning, statistical learning theory (SLT). 
Our discussion is mainly based on the textbooks by \citet{ShaBen14} and \citet{HarRec22}. We will prioritize conceptual insights over completeness.%
    \footnote{An introduction to SLT for philosophers is \citep{LuxSch11inc}. A more basic philosophical introduction is \citep{HarKul07}, and a shorter summary is given in \citep{Ste26philsci}.}

\subsection{The problem of generalization}

The most basic type of machine learning problem is classification. Consider the CIFAR-10 database, consisting of 60,000 images of 32x32 pixels. Each image is in one of ten different classes (airplanes, cars, birds, cats, deer, dogs, frogs, horses, ships, and trucks). A machine learning algorithm for classification is trained on (part of) this dataset, with the aim of producing a model which \emph{generalizes}: which correctly classifies new and unlabeled images.

In SLT, problems of this type are formalized as follows. We have a domain $\mathcal{X}$ of \emph{instances}, which are usually themselves vectors of real-valued \emph{attributes}; in the CIFAR-10 example, the instances would be the images, constituted by color values for each pixel. We have a \emph{label} set $\mathcal{Y}$; in the example, these are the ten possible classes. A \emph{classifier} (also 
 \emph{model}) is a function $h: \mathcal{X} \rightarrow \mathcal{Y}$ from all possible instances to labels. A \emph{learning algorithm} $A$ is a function that receives a training \emph{sample} $S$, a finite ordered sequence of instance-label pairs, and returns a classifier. 

To give an even simpler example than CIFAR-10 (inspired by \citealp{ShaBen14}), consider the problem of predicting whether a mango on display at the Groningen market is good from just two real-valued features, its color (ranging from 0, dark green, to 1, dark brown) and its softness (ranging from 0, rock hard, to 1, mushy), based on a labeled sample of mangoes you bought and tried earlier. This is a binary classification problem: there are just two possible labels (good or not, 0 or 1). The instances are points in the square $[0,1]^2$. Any classifier simply selects a subset of this square. For instance, a linear separator can be visualized as a line dividing the square into two, with the instances falling on one side of the line classified as good.

A crucial assumption in SLT is that training instances as well as new data instances are independently and identically distributed (i.i.d.)\ samples from some true but unknown distribution $\mathcal{D}$ over $\mathcal{X} \times \mathcal{Y}$.\footnote{This
    is a substantial ``uniformity of nature'' assumption. But it is not an assumption that is at stake in the generalization puzzle.} 
This assumption allows us to define, for any given classifier $h$, the probability that it misclassifies a randomly picked instance, 
\begin{align}
L_\mathcal{D}(h) := \mathbb{P}_{(X,Y) \sim \mathcal{D}}\left[h(X) \neq Y \right].    
\end{align}
We call this probability the \emph{true risk} of $h$, and the goal of learning is to find a classifier as close as possible to the so-called Bayes optimal classifier, which minimizes the true risk.

Since in a learning problem we do not know the true distribution, we cannot simply use an algorithm to calculate true risks and determine the Bayes optimal classifier. Instead, what a learning algorithm \textit{can} calculate is the performance of a given classifier on the labeled training sample. The \emph{empirical error} of classifier $h$ on sample $S$ is its mean number of misclassifications,
\begin{align}
L_S(h) := \frac{|\{(x,y)\in S: h(x) \neq y \}|}{|S|}.
\end{align}
Thus a learning algorithm can in principle find a classifier that is successful on the training data (has low empirical error); but we would only say that the algorithm has learned successfully if the selected classifier also {generalizes} well to \textit{unseen} data (has low true risk). This leads to the problem of generalization, and the approach in SLT is to determine under which conditions empirical error is indeed a good indication of true risk: what you see is what you get (``wysiwyg,'' \citealp{Bel21ac}). 

\subsection{The generalization gap and model complexity}\label{ssec:erm}
 
The quantity at the center of SLT is the \emph{generalization gap} between what you see and what you get,
\begin{align}\label{eq:gengap}
\Delta_{A,S} := \left| L_\mathcal{D}(A(S)) - L_S(A(S)) \right|,
\end{align}
where $\hat{h}=A(S)$ is the classifier returned by learning algorithm $A$ when trained on sample $S$. A natural choice of learning algorithm $A$ is one that, for given training data $S$, selects a classifier $\hat{h}$ that minimizes the empirical error $L_S(\hat{h})$. This is Vapnik's \citeyearpar{Vap00} first ``inductive principle,'' or the learning rule of \emph{empirical risk minimization} (ERM). A bound on the generalization gap \eqref{eq:gengap} for this method would give us, in virtue of minimal empirical error, a near-minimal true risk.

\subsubsection{Representation and optimization}

To implement ERM as a procedure we need to make at least two further choices: which classifiers to consider in the first place, and how to break ties among equally good classifiers. 

We must make the first choice because there always trivially exist 
any number of 
classifiers that fit given data perfectly but behave differently on unseen instances. 
In reality, machine learning algorithms work with a restricted \emph{model class} of classifiers. The definition of ERM also presupposes some such choice of model class $\mathcal{H}$,
\begin{align}\label{eq:erm}
\mathrm{ERM}_\mathcal{H}(S) \in \argmin_{h \in \mathcal{H}}L_S(h).
\end{align}

The choice of model class is the
problem of \emph{representation}. In the case of a neural network, the model class is the class of all functions expressible by some setting of connection weights given the architecture: each setting of weights determines a classifier, and learning consists in finding a good setting. The class of linear separators, the larger class of quadratic separators, and the class of all polynomial separators are other examples with increasing expressiveness.

We must make the second choice because there might still be multiple classifiers in chosen $\mathcal{H}$ which have the same minimal empirical error on data $S$, and rule \eqref{eq:erm} does not specify how such ties are broken. 
This is part of 
the problem of \emph{optimization},
which concerns how to actually implement the ERM algorithm.
In deep learning, the standard approach is a version of stochastic gradient descent (SGD). Notice that optimization and representation are linked, because the actual learning or optimization algorithm may 
only ever converge on a strict subset of $\mathcal{H}$, so that the \emph{effective capacity} may be smaller than that of the nominal model class.

In our exposition of the classical theory we set optimization aside and analyze ERM directly, since for the relevant theoretical results only property \eqref{eq:erm} matters. The trichotomy of generalization, representation, and optimization will each reemerge in our later discussion of benign interpolation.

\subsubsection{Uniform convergence and model class capacity}
Statistical learning theory makes precise what we need to assume about $\mathcal{H}$ in order to obtain a bound on the generalization gap. The central result is a \emph{uniform convergence} bound, and the core property of the model class is a notion of its complexity or \emph{capacity}.
 
A training sample $S$ is $\epsilon$-\emph{representative}, for model class $\mathcal{H}$ and accuracy parameter $\epsilon>0$, if simultaneously for all classifiers $h \in \mathcal{H}$ the difference between $h$'s empirical error $L_S(h)$ on $S$ and $h$'s true risk $L_\mathcal{D}(h)$ is smaller than $\epsilon$:
\begin{align}
(\forall h \in \mathcal{H}) \left[  | L_S(h) - L_\mathcal{D}(h)| \leq \epsilon \right].
\end{align}
On such a sample, what-you-see-is-what-you-get (up to $\epsilon$). The class $\mathcal{H}$ has the \emph{uniform convergence property} if for any chosen confidence parameter $\delta>0$, there is some training sample size $m$ (dependent on $\epsilon$ and $\delta$) such that for \emph{any} unknown distribution $\mathcal{D}$ over samples $S$,
\begin{align}\label{eq:unifconv}
  \textrm{Prob}_{S \sim \mathcal{D}^m}\left[(\forall h \in \mathcal{H}) \left[  | L_S(h) - L_\mathcal{D}(h)| \leq \epsilon \right] \right] \geq 1-\delta.  
\end{align}
Since this holds for all classifiers in $\mathcal{H}$ simultaneously, it holds for any classifier selected by our learning algorithm $A$: with high probability and for large enough training sample, we have an $\epsilon$-bound on the generalization gap \eqref{eq:gengap}. 
 
In particular, if a model class has the uniform convergence property, then this justifies the ERM rule, since with high probability, in virtue of selecting a classifier with minimal empirical error, it selects one with near-minimal true risk. We say that ERM$_\mathcal{H}$ \emph{learns} the model class $\mathcal{H}$: for any chosen $\epsilon, \delta > 0$, for large enough $m$ and for any $\mathcal{D}$,
\begin{align}
  \textrm{Prob}_{S \sim \mathcal{D}^m}\left[  L_\mathcal{D}(\mathrm{ERM}_\mathcal{H}(S)) - \min_{h \in \mathcal{H}} L_\mathcal{D}(h) \leq \epsilon \right] \geq 1-\delta.  
\end{align}
Beyond this probabilistic guarantee of \emph{reliability}, the wysiwyg property \eqref{eq:unifconv} is important in itself, because it tells us that the training error of the selected model is probably a good indication of its true risk: we will probably see it if the selected model (and therefore chosen model class) is bad.
 
What kind of model classes have the uniform convergence property? To answer this we need to consider the \textit{capacity} of a model class.\footnote{In the best studied setting of binary classification, the relevant capacity notion is the Vapnik-Chervonenkis (VC) dimension (see \citealp[ch.\ 6]{ShaBen14}); for multiple classes (like the CIFAR-10 example), there is a generalization called Natarajan dimension (see \citealp[ch.\ 29]{ShaBen14}). More formally, the \emph{restriction of $\mathcal{H}$ to finite set $X$} is the class $\mathcal{H}_{|X}$ of functions $f: X \rightarrow \mathcal{Y}$ such that $f(x) = h(x)$ for some $h \in \mathcal{H}$ and all $x \in \mathcal{X}$. Then $\mathcal{H}$ shatters finite $X \subset \mathcal{X}$ if the restriction of $\mathcal{H}$ to $X$ contains \emph{all} functions $f: X \rightarrow \mathcal{Y}$, that is, $|\mathcal{H}_{|X}|=2^{|X|}$. The VC dimension of $\mathcal{H}$ is the maximal size of a set $X \subset \mathcal{X}$ that is shattered by $\mathcal{H}$.} Roughly, capacity measures how flexible a class is in fitting any possible training sample, and as such it is a measure of the richness or complexity of the class; a model class with lower capacity is in that sense \emph{simpler}.\footnote{See \citep[sect.\ 3]{Ste25mam} for further discussion of this interpretation.} For instance, the class of linear separators has strictly lower capacity than the class of quadratic separators, while the class of all polynomial separators has infinite capacity. In the case of neural nets, capacity is related to the size of the network, in particular the number of weights \citep{AntBar99}.  
 
The fundamental result of SLT is that a model class has the uniform convergence property if and only if its capacity is finite.\footnote{This is the ``fundamental theorem of statistical learning theory,'' due to \citet{VapChe71tpa}. See, e.g., \citep[ch.\ 6]{ShaBen14}.} A quantitative version of the result says further that the uniform convergence property is stronger when the capacity is smaller. We thus have stronger probabilistic  wysiwyg and reliability guarantees when the model class is simpler. This underwrites a methodological simplicity norm:\footnote{See \citep{Ste25mam} for a detailed exposition of this ``core argument'' for a simplicity preference.}
 
\begin{quote}
\textbf{Methodological norm \customlabel{label:m1}{M1} (Occam's razor).} In order for a learning algorithm to have a probabilistic guarantee of good generalization, its model class must be simple.
\end{quote}

By ``good generalization'' we mean here (and elsewhere) a small generalization gap. As we will explain, this norm expresses the familiar intuition from statistical practice that a model class that is not simple will likely \emph{overfit}. 
But it leaves something out: the risk of \emph{underfit}.

\subsection{The bias-complexity trade-off and explicit regularization}
\label{ssec:srm}
 
The generalization gap \eqref{eq:gengap} only concerns the difference between a classifier's empirical error and true risk: it does not say anything about how good these errors are in an \textit{absolute} sense. Learnability of $\mathcal{H}$ by ERM$_\mathcal{H}$ is a probabilistic guarantee of finding the near-best classifier in $\mathcal{H}$, but the best classifier in $\mathcal{H}$ might still be bad. The theoretical analysis is concerned with the \emph{estimation error}
\begin{align}
\epsilon_{\mathrm{est}} := L_\mathcal{D}(A(S)) - \min_{h \in \mathcal{H}} L_\mathcal{D}(h),
\end{align}
but not directly with the \emph{approximation error}
\begin{align}
\epsilon_{\mathrm{app}} := \min_{h \in \mathcal{H}} L_\mathcal{D}(h),
\end{align}
which expresses how good the model class $\mathcal{H}$ was to begin with. 
 
\subsubsection{A check against simplicity}

The choice of model class amounts to introducing an \emph{inductive bias}: a restriction to some set of classifiers that is reasonable for the task at hand.\footnote{More specifically, this is a \emph{restriction} inductive bias, as opposed to a \emph{preference} inductive bias  \citep[p.\ 64]{Mit97}. That any learning algorithm must operate with an inductive bias is the lesson of the so-called no-free-lunch theorems (see \citealp{SteGru21syn}); also see \S \ref{ssec:thegap} below.} A bad choice results in high approximation error.
 
The simplicity norm \ref{label:m1} tells us to strive for a simple model class; but we do not want to make the class \emph{too} simple, if that means that even the best classifier in the class has high true risk. The push towards a simple $\mathcal{H}$ must therefore be checked by a more informal assessment of whether the inductive bias represented by $\mathcal{H}$ is still in line with what we believe or are willing to assume about the learning problem. This makes \ref{label:m1} unsatisfying in two related ways. We cannot follow it in situations where we are unwilling or unable to make sufficiently strong assumptions, for instance because we know too little about the domain; and such situations are characteristic of machine learning, which is usually contrasted with traditional statistical inference as a primarily data-driven approach, with minimal modeling assumptions.

\subsubsection{The bias-complexity trade-off}
\begin{figure}
\begin{center}
\begin{tikzpicture}
  \begin{axis}[
      width=10cm,
      height=6cm,
      axis lines=middle,
      xlabel={capacity of $\mathcal{H}$},
      xlabel style={yshift=-1.8em},
      xtick=\empty,                
      ytick=\empty, 
      xmin=0, xmax=10,
      ymin=0, ymax=6,
      domain=0:10,
      tick label style={font=\tiny,opacity=0},
    ]
    \addplot[thick] {0.15*(x-5)^2 + 1.7}
      node[pos=0.35,anchor=south,yshift=3mm] {\small true risk};
    \addplot[thick, dashed] {5.25 * 1.35^(-x - 0.1 * x^2)+0.2}
      node[pos=0.60,anchor=south,yshift=2mm,xshift=1mm] {\small appr.\ error};
    \pgfmathsetmacro{\yVar}{0.4}
    \pgfmathsetmacro{\yTot}{3.05}
    \draw[densely dashed,very thick] (axis cs:8,\yVar) -- (axis cs:8,\yTot);
    \node[rotate=90,above] at (axis cs:8,{(\yVar+\yTot)/2}) {\small est.\ error};
  \end{axis}
\end{tikzpicture}
\caption{The bias-complexity trade-off.}\label{fig:biasvar}
\end{center}
\end{figure}
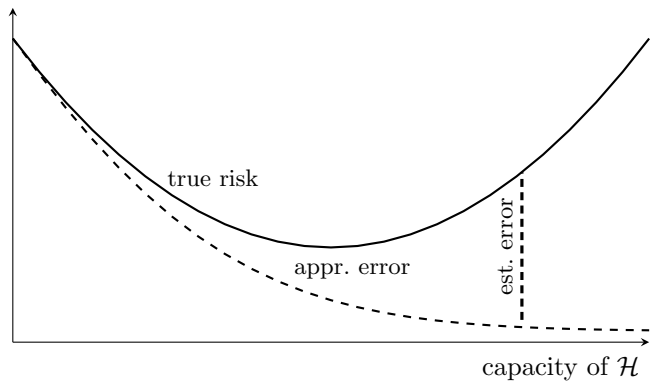

Zooming out, we see the minimization of estimation error and the minimization of approximation error as pulling in different directions. There is a \emph{bias-complexity trade-off} 
(figure \ref{fig:biasvar}).
 
With a model class of minimal capacity, approximation error can be expected to be high: the class may \emph{underfit}. As we increase capacity, we may hope to include better classifiers. However, as the class gets too complex, it will \emph{overfit}: it will fit the training data very well, but not generalize as well. There is asymmetry in how the two sides can be analyzed. Low approximation error is ultimately a matter of choosing a good model class, which is to some extent a matter of what we know about the domain, and the analysis here remains informal. Controlling estimation error, in contrast, has the formal backing of the theory of uniform convergence. What we show next is that the trade-off can itself be partly formalized, leading to a second and apparently less stringent simplicity norm.

\subsubsection{Generalized uniform convergence and explicit regularization}
We can consider a generalized version of the uniform convergence property.\footnote{See \cite{Ste26philsci} for a more detailed account of the following reasoning.} Instead of a single class of finite capacity, consider a sequence $(\mathcal{H}_i)_{i \in \mathbf{N}}$ of such classes, possibly constituting a very complex superclass $\mathcal{H}= \cup_{i} \mathcal{H}_i$. 
Each $\mathcal{H}_i$ has the uniform convergence property, so for each we have a wysiwyg bound: for given sample size $m$ and confidence $\delta$, with probability at least $1-\delta$,
\begin{align}\label{eq:unifconv2}
  (\forall h \in \mathcal{H}_i) \left[  | L_S(h) - L_\mathcal{D}(h)| \leq \epsilon_i(m,\delta) \right],
\end{align}
where $\epsilon_i(m,\delta)$ depends on the capacity of $\mathcal{H}_i$, getting weaker as capacity increases.
 
It can be shown that, \emph{for all $\mathcal{H}_i$ simultaneously}, we have (with probability at least $1-\delta$)
\begin{align}\label{eq:genunifconv}
  (\forall i \leq n)(\forall h \in \mathcal{H}_i) \left[  | L_S(h) - L_\mathcal{D}(h)| \leq \epsilon_i(m,\delta/n) \right].
\end{align}
The price we pay for this uniform bound is a looser $\epsilon_i(m,\delta/n)$ for each individual $\mathcal{H}_i$; what we gain is a single bound that holds across all classes. Now, just as ERM arises from explicitly minimizing the uniform convergence bound \eqref{eq:unifconv2} (since the accuracy term is constant within a fixed class, the classifier with smallest empirical error gives the sharpest bound on true risk), this generalized bound underlies Vapnik's second ``inductive principle,'' the rule of \emph{structural risk minimization} (SRM), 
\begin{align}
\mathrm{SRM}_\mathcal{H}(S) \in \argmin_{h \in \cup_{i \leq n} \mathcal{H}_i}  L_S(h)+ \epsilon_{i(h)}(m,\delta/n),
\end{align}
where $i(h)$ denotes the first $i$ such that $h \in \mathcal{H}_i$. SRM thus selects a classifier that minimizes the sum of empirical error and a penalty term. Since this term depends only on $\mathcal{H}_{i(h)}$, and is larger the higher the capacity of that class, it functions as a penalty for model-class complexity. By trading empirical fit for complexity, SRM implements a form of explicit \emph{regularization}.\footnote{This is a type of \emph{preference} inductive bias: for equal fit, models from the simpler model class are preferred.}

 
Analogously to the ERM case, these results give a reliability and wysiwyg justification for SRM and its methodology of regularization, underwriting a more refined methodological norm:\footnote{  By ``good generalization'' we again mean a small generalization gap, so small overfit. Note, however, that this probabilistic guarantee is now subclass-dependent: it is stronger for lower-capacity subclasses. Spelling out the justification for this methodological norm involves some subtleties, including the need for a pragmatic ``luckiness'' reasoning. See again \citep{Ste26philsci} for details.}
 
\begin{quote}
\textbf{Methodological norm \customlabel{label:m2}{M2} (Occam's razor).} In order for a learning algorithm with a complex model class to have a probabilistic subclass-dependent guarantee of good generalization, it must trade fit for simplicity.
\end{quote}
 
A suite of methods for statistical model assessment and selection can be traced back to this norm, or pair of norms \citep{ClaeskensHjort2012}. Trade-offs of fit for simplicity show up in various information criteria and in Bayesian model comparisons. This convergence is not surprising, since these methods all ultimately derive from the goal of accurate prediction. Insights from this literature have also made their way into philosophy of science (see \citealp[ch.\ 2]{Sob15}), 
supporting an empiricist reading of Occam's simplicity norm: we trade fit for simplicity not because this matches how we imagine the world to be, but because we thereby steer a course between two kinds of prediction error. If we make the model too simple we fail to detect relevant patterns, and underfit; if we make it too complex we identify patterns where there is only noise, and overfit.

Specifically, as argued in detail by \citet{Ste26philsci}, and against some persistent views in the philosophy of science (e.g., \citealp[chs.\ 6--7]{Nor21}, \citealp{BarCevGne22mam}), the two norms are \emph{not} dependent on a material assumption on the world (or, more modestly, on the relevant domain) that simpler models or model classes are (more likely to be) more accurate. In short, the reasoning underwriting these norms does not use such assumptions; and so these norms remain valid even if we do not want to make such assumptions. As we will see, this stands in stark contrast to the form of Occam's razor that emerges from work on explaining benign overfitting.
 
\subsubsection{The theory and the practice}
It is still a jump from the above theory to the actual practice of machine learning. As also discussed by \citet[p.\ 111]{GooBenCou16}, the generalization bounds central to our discussion are not directly used in the design of learning algorithms: they are very worst-case and therefore loose, and in the case of deep learning it is not easy to determine the precise capacity of a model class. Nevertheless, these bounds still provide an ``intellectual justification'' \citep[p.\ 111]{GooBenCou16} for standard approaches. The two methodological simplicity principles we identified are (or at least, \emph{were}) core to machine learning practice. While learning-algorithm designers may not seek directly to implement SRM, it has been normal practice to follow \ref{label:m2} and employ some form of explicit regularization. These two norms have, however, recently come under pressure: it is recognized as a genuine issue in the field that they appear to have lost some of their prescriptive as well as explanatory value.

\subsection{The object of simplicity}
\label{sec:classsimplicity}

Before turning to that pressure, it is worth dwelling on the fact that this notion of simplicity attaches to the model class and not to the individual classifiers in it. The two come apart, and in more than one way. On the usual ways one would try to define the simplicity of individual models (for instance, in terms of description length, or number of parameters), there will be fewer simpler than more complex models, and so a class of simple models will be small and so have low capacity \citep[fn.\ 14]{Ste26philsci}; 
but the converse fails. One can assemble a small, low-capacity class entirely out of individually complex (on any of these notions) models, and the guarantees above apply to it unchanged \citep[sect.\ 3.1]{Ste25mam}. What the fundamental theorem responds to is the capacity of the class, not the nature of its members: as \citet{Pea78} observed of the capacity-based arguments, the appeal to simplicity of the members is ``only incidental'' (p. 263). 

The simplicity of an individual hypothesis is not a stable property to begin with. Description length depends on the encoding, and an encoding that makes one hypothesis short can be exchanged for one that makes it long, so there is no encoding-invariant fact about which models are simple (\citealp{Her20pos}, sect.\ 5.1; \citealp{Dom99dmkd}). 
Similarly, to describe a family of functions by a particular set of parameters is itself to choose an encoding, with no more claim to privilege than any other; which is why parameter count need not track capacity. The one-parameter family of sine curves $\{x \mapsto \sin \alpha x\}_{\alpha \in \mathbb{R}}$ is as parsimonious as a parametrization could be, yet has infinite VC dimension \citep[p.\ 78]{Vap00}. The single parameter is one description among many, and nothing makes it answer to the class's ``inherent complexity'' or how flexibly it can fit data (\citealp{Rom17pos}; \citealp[pp.783f]{Kie01bjps}).

This is the situation against which the achievement of the classical theory should be read. ``Prefer the simpler model'' has no determinate content on its
own: even if we were to agree on one of the candidate notions of individual simplicity, it is only fixed relative to 
a ``language,'' an encoding or a parametrization, that the structure of a learning problem leaves open. As such, the prospects for a robust formal connection of any such notion to good generalization look slim; and we indeed know of no such results. 
The classical theory 
instead moves to a property of the class, capacity, which is a more robust notion and, crucially, one that is provably connected to 
generalization. The fundamental theorem is what 
gives the preference for simplicity a definite sense. This is why, as we will see, the move in the new accounts back to a notion of simplicity that attaches to individual models is more consequential than it looks. 

\section{The generalization puzzle}
\label{sec:Generalization}
 
With the standard treatment of machine learning methods in place, we are ready to introduce the puzzle that is central to this paper. How is it that the perfect fit to data achieved in machine learning methods is benign and yet does not lead to overfitting? In this section and the next we review parts of the fast-moving statistics and computer science literature on this issue, explaining it for a philosophical audience up to the point where we can develop our central claims.
 
\subsection{The puzzle}
 
We begin with two striking observations from \citet{ZBHRV17iclr}.\footnote{Their work was reprinted and updated as \citep{ZBHRV21acm}.}
 
\subsubsection{Fitting random labels}\label{sssec:zhangetal}
 
Here is the first observation. \citet{ZBHRV17iclr} take a standard image-recognition dataset and replace the labels of the training instances with \emph{randomly generated} labels. Since the true relationship between instances and labels is now fully random, there is nothing to learn. Nevertheless, popular deep learning networks ``\emph{easily fit random labels}'' (ibid., p.\ 2): the algorithm achieves near-perfect training error on this meaningless data. At the same time, on the original data with non-random labels, the same networks achieve both low training error and 
(as a proxy for true risk) 
low test error, as was well known. The combined picture, summarized in table \ref{tb:random}, is what poses the puzzle.\footnote{Where we here and in the following speak about generalization error, strictly speaking (since test error is an estimate of true risk) we mean the corresponding estimate of generalization error.}
 
\begin{table}[h!]
\begin{center}
\begin{tabular}{ c|c|c|c } 
  & train error & test error & gener.\ error \\ \hline
 original labels & low & low & low \\ \hline
 random labels & low & high & high \\  
\end{tabular}
\end{center}
\caption{The randomization experiment of \citet{ZBHRV17iclr}.}\label{tb:random}
\end{table}
 
Why is this puzzling? The relevant deep learning architectures have a number of free parameters large enough to wildly exceed the number of data instances. This \emph{overparametrization} suggests that the capacity of the model class is excessive. One natural response, however, is to distinguish the nominal capacity of the architecture from its \emph{effective capacity}, ``the size of the subset of models that is effectively achievable by the learning procedure'' (\citealp[p.\ 109]{ZBHRV21acm}; also see \citealp[pp.\ 110f]{GooBenCou16}). Due to computational constraints or specific properties of the optimization algorithm, the models that the algorithm can actually reach may be much more limited than those that the architecture could in principle express. If the effective model class is sufficiently simple (and has the right inductive bias for image recognition), the classical story still predicts what we see on the natural data: low training error and low test error.
 
The random-label results throw cold water on this resolution. If the effective model class were simple, the classical story predicts that the generalization gap would be small, so that high true risk (forced by the random labels) would have to manifest as high training error. This, however, is not what Zhang et al.\ observe: the random labels are fitted nearly perfectly, while test error is high. Therefore, what you see is \emph{not} what you get. The experiment implies that deep neural nets do \emph{not} have a simple model class, even in the effective sense---yet they generalize well on natural data.\footnote{As before and after, by ``good generalization'' we mean a small generalization gap: good training error and good true risk (test error), so no overfitting. }\footnote{\label{fn:neysha}\citet{NeyTomSre15iclr} made related observations when training networks of increasing size on the MNIST and CIFAR-10 datasets, noting ``the test error continues decreasing'' past the point needed to achieve zero training error, a behavior ``not at all predicted by, and even contrary to, viewing learning as fitting a model class controlled by network size'' (p. 2). They experiment with partly random data as well, and propose an analysis in terms of ``implicit regularization,'' (p. 3) though this is explicitly in terms of ``capacity control'' (p. 1).}
 
\begin{quote}
\textbf{Observation.} Our learning algorithm does not have a simple model class, yet generalizes well.
\end{quote}

\subsubsection{The role of regularization}\label{sssec:regular}
 
The second main observation of \citet{ZBHRV17iclr} concerns regularization. In line with our discussion in §\ref{ssec:srm}, they write that ``regularization can be thought of as the operational counterpart of a notion of model complexity [\dots]\ regularization introduces algorithmic tweaks intended to reward models of lower complexity'' \citeyearpar[p.\ 108]{ZBHRV21acm}. Even if the model class has high 
capacity, various regularization techniques can ``confine learning to a subset of the [model] space with manageable complexity'' \citeyearpar[p.\ 6]{ZBHRV17iclr}. Zhang et al.\ investigate what happens when three popular regularizers for neural nets---data augmentation, weight decay, and dropout---are switched on and off. Their finding is that ``\textit{regularization may improve generalization performance, but is neither necessary nor by itself sufficient for controlling generalization error}'' (ibid., p.\ 2). Even with all regularizers off, the learning algorithm achieves (slightly worse but still) very good test accuracy on natural data. This indicates that regularization is not necessary for good generalization.\footnote{Even when one or more of these regularizers are turned on, they find that in most instances the learning algorithm still achieves (not necessarily near-perfect but still) extremely good fit on the randomly generated data. This shows that these regularizers are not sufficient for good generalization either; suggesting they are not actually regularizing enough.}
 
\citet{BelMaMan18icml} generalize the point beyond deep neural nets, examining \emph{kernel machines}. 
As also discussed by \citet{Bel21ac}, 
they show that kernel machines can similarly achieve zero training error, and argue that explanations based on classical generalization bounds, including 
bounds for regularizers, are ``implausible, if not outright impossible'' (ibid., p.\ 211).
 
The essential point is that 
wysiwyg bounds are stretched into vacuity for models that perfectly fit noisy data.\footnote{The basic reasoning applies to the findings of \citet{ZBHRV17iclr} as well, though \citet[pp.\ 210--11]{Bel21ac} remarks that this does not yet rule out application of classical data-dependent bounds, like margin bounds.} To make this concrete, consider a true distribution $\mathcal{D}$ that is noisy, so that for the Bayes optimal model $h^*$ we have $L_\mathcal{D}(h^*)=q>0$. Then since for interpolating (perfect-fit) model $\hat{h}$ we have $L_S(\hat{h})=0$, the generalization gap is
\begin{align}
    L_\mathcal{D}(\hat{h}) - L_S(\hat{h}) = L_\mathcal{D}(\hat{h}) \geq L_\mathcal{D}(h^*)=q.
\end{align}
No bound on the generalization gap could be sharper than the constant noise level $q$, no matter how large the quantity of training data, and so such bounds would be vacuous.\footnote{Also see \citet[sect.\ 2.9]{BarMonRak21ac} for a more detailed explanation of the ``mismatch between benign interpolation and uniform convergence'' for neural nets.} Conversely, given that there can be no non-vacuous bounds for the learning algorithm in question, it cannot be the case that this is an explicit regularizer for which such bounds would be derivable. In short:
 
\begin{quote}
\textbf{Observation.} Our learning algorithm with a complex model class does not trade fit for simplicity, yet generalizes well.
\end{quote}

\subsubsection{Benign interpolation and double descent}
 
Modern machine learning thus fails to conform to the statistical intuition, backed up by the classical theory, that a perfect fit of training data indicates an overly complex model class and likely overfitting.\footnote{Although
    it has later been noted that such phenomena had already been observed decades prior \citep{LVMKT20nas}.} 
The phenomenon has been named \emph{benign overfitting} \citep{BarLonLugTsi20pnas}; but as \citet[pp.\ 5f]{Cur25arx} notes, ``overfitting'' already connotes bad generalization, so that ``benign overfitting'' sounds like a contradiction. Following Curth, we adopt the term \emph{benign interpolation}.\footnote{Another term is ``harmless interpolation'' \citep{MutVodSubSah20ieee}.}
 
\begin{figure}
\begin{center}
\begin{tikzpicture}
  \begin{axis}[
      width=12cm,
      height=7cm,
      axis lines=middle,
      xlabel={capacity of $\mathcal{H}$},
      xlabel style={yshift=-3.5em},
      xtick=\empty,
      ytick=\empty,
      xmin=0, xmax=12,
      ymin=-0.8, ymax=6,
      tick label style={font=\tiny,opacity=0},
      clip=false,
    ]
    \addplot[thick, no markers, domain=0.6:6, samples=200] {0.3571*(x-3.2)^2 + 1.4};
    \addplot[thick, no markers, domain=6:11.7, samples=200] {0.15 + 4.05*exp(-0.65*(x-6))};
    \draw[densely dashed] (axis cs:6,0) -- (axis cs:6,4.3);
    \node[anchor=west, font=\small] at (axis cs:3.9,5.2) {interpolation threshold};
    \node[font=\small] at (axis cs:3,-0.5) {classical regime};
    \node[font=\small] at (axis cs:9,-0.5) {modern regime};
    \node[anchor=south, font=\small] at (axis cs:1.4,4.2) {test error};
  \end{axis}
\end{tikzpicture}
\caption{The double-descent curve. In the classical regime, test error follows the u-curve of the bias-complexity trade-off. Past the interpolation threshold, where the model class is large enough to fit the training data perfectly, test error descends again, even dropping below the classical minimum.
}\label{fig:doubledescent}
\end{center}
\end{figure}
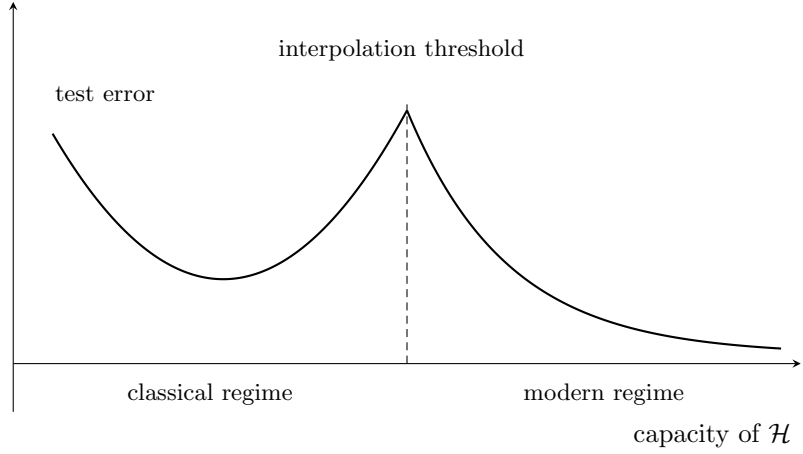

Perhaps most influential in the subsequent literature has been the reappraisal by \citet{BHMM19pnas} of the classical u-curve of the bias-complexity trade-off (recall figure \ref{fig:biasvar}). They note that ``[c]onventional wisdom in machine learning suggests controlling the capacity of the function class $\mathcal{H}$ based on the bias-variance trade-off [\dots]\ classical thinking is concerned with finding the `sweet spot' between underfitting and overfitting'' (ibid., p.\ 15849). But the new ``best practice'' for choosing neural network architectures, namely architectures ``large enough to permit effortless zero-loss training,'' runs against this classical picture (ibid.). For a number of learning procedures---neural networks, decision trees, ensemble methods---Belkin et al.\ investigate performance for model classes of increasing capacity. What they observe is that before the ``interpolation threshold,'' where model capacity is just large enough to fit the training data perfectly, test error first drops and then rises again, in line with the classical picture. When pushing capacity further, however, test error decreases \emph{again}. After the initial classical descent and ascent, we observe a second descent: the u-curve extends to a ``double-descent'' curve (figure \ref{fig:doubledescent}).\footnote{\citep{CurJefSch23neurips} 
    criticize this analysis for several of the learning procedures, arguing that on the right way of measuring capacity (by effective number of parameters) performance ``folds back'' into the classical u-shape. However, they do not address the case of deep learning.}

\subsection{The challenge}
 
\citet[p.\ 10]{ZBHRV17iclr} write that their observations pose a ``conceptual challenge to statistical learning theory as traditional measures of model complexity struggle to explain the generalization ability of large artificial neural networks.'' What exactly is this challenge?
 
\subsubsection{The status of the classical theory}
Benign interpolation is not \emph{inconsistent} with the classical theory: it could not be, since SLT is a body of mathematical results. It is rather that, in cases of benign interpolation, the mathematical antecedents are not satisfied---the model class is not simple, and the algorithm does not trade simplicity for fit. Strictly speaking, the methodological norms \ref{label:m1} and \ref{label:m2} remain intact: SLT still tells us that in order to benefit from its probabilistic guarantees we must keep the model class simple or have our algorithm trade fit for simplicity.\footnote{The
    theory does not strictly speaking tell us that if our algorithm does \emph{not} have a simple model class or does \emph{not} trade fit for simplicity it will \emph{not} generalize well, even if this is the classical statistical intuition \citep[p.\ 221]{HasTibFri09}. To be more precise, the fundamental theorem says that a simple class is necessary and sufficient for a probabilistic \emph{guarantee} of successful generalization; but of course successful generalization is possible absent such a guarantee. In particular, the guarantee is worst-case over all possible distributions, leaving open that a particular complex model class works well for a restricted collection of matching distributions. Here we already see room for an answer to the challenge (cf.\ \citealp{BatWoo25arx}; also see \S\ref{sssec:retreat}).}
But the proliferation of cases where algorithms generalize well without heeding these norms reveals the limited applicability of the classical theory and puts pressure on its prescriptive value.\footnote{There 
    are still many scenarios in modern machine learning where interpolation does \emph{not} lead to good generalization, and a classical account still applies (cf.\ \citealp[pp.\ 205-06]{Bel21ac}; also see §\ref{sssec:retreat}).  But the phenomenon is sufficiently common and striking to make the apparent inapplicability of the classical theory an important and interesting problem.} 
Both ``the prescriptive and descriptive value of these theories [i.e., classical guarantees] remains debated'' \citep[p.\ 107]{ZBHRV21acm}.

The more common framing, in the literature we are reviewing, is that the shortcoming is explanatory. In the passage quoted above, \citet{ZBHRV17iclr} link their challenge to SLT's struggle to \emph{explain} generalization in cases of benign interpolation. Similarly, \citet{BelMaMan18icml} write that ``existing bounds seem to provide little explanatory power,'' and \citet[p.\ 112]{BarMonRak21ac} write that ``mechanisms of uniform convergence alone cannot explain good statistical performance of [interpolating] methods.''\footnote{\citet{Bel21ac} also uses the vocabulary of explanation throughout.}
 
\subsubsection{Explaining generalization}
\label{subsec:explaining}
 
What is at stake, therefore, is not the classical normative principle but an explanatory principle embedded in the classical theory.
 
\begin{quote}
\textbf{Explanation (classical).} Our learning algorithm generalizes well, because it has a simple model class or trades fit for simplicity. 
\end{quote}
 
Since, for cases of benign interpolation, our learning algorithm does \emph{not} have a simple model class, \emph{nor} does it trade fit for simplicity, yet it generalizes well, this explanation does not apply. 
 
\begin{quote}
\textbf{Explanatory challenge.} Why does our learning algorithm generalize well, given that it neither has a simple model class nor trades fit for simplicity?
\end{quote}

It is natural to perceive this as an issue of the theory lagging behind the practice of machine learning, so that the challenge is to develop an improved mathematical theory. 
But an improved theory of generalization, or a ``new framework for a `theory of induction{'}'' \citep[p.\ 217]{Bel21ac}, need not be a purely mathematical framework. For instance, \citet[p.\ 123]{HarRec22} pose as an open question, ``What is it a successful theory of generalization should do?'', and note that ``[e]ven a qualitative theory of generalization may be useful''. When we consider the main theoretical approaches offered to explain benign interpolation, we will return to the question of what kind of explanation of generalization this emerging theory offers.\footnote{Explaining generalization is only one aspect of making sense of deep learning. Much work in philosophy and computer science is concerned with understanding trained deep learning models in light of their \emph{opacity} (see, e.g., \citealp{BeiRaz22pc}). Two kinds of opacity are usually distinguished: opacity about what a trained model has learned (``w-opacity,'' \citealp{Bog22mam}; or ``inference-opacity,'' \citealp{Sog23cjop}) and opacity about how the model came to learn it (``h-opacity'' or ``training-opacity''). The latter connects more closely to the problem of generalization, since understanding why a training process yields models that generalize is part of understanding what that process does. We do not argue here that understanding generalization is necessary for understanding deep learning models (see \citealp[sect.\ 5.2]{RazBei24erk} for such a claim), but focus on it as a question of independent interest.}

\section{Implicit inductive bias and Occam's razor}
\label{sec:implicit}
 
Having set out the explanatory challenge posed by benign interpolation, we now turn to attempts by computer scientists to meet it. A common dialectical structure runs through what we call the \emph{basic strategy}: it locates the explanation of benign interpolation in the learning algorithm's \emph{implicit inductive bias} toward simple models. As we lay out the strategy in three steps, it will be important to keep in view a shift that distinguishes it from the classical story. The simplicity preference that SLT underwrites (\ref{label:m1} and \ref{label:m2}) is a preference for \emph{model classes} of low capacity. The simplicity preference at work in the basic strategy is a preference over \emph{individual models}. These are different targets. The classical theory gives us a provable connection between the simplicity of a model class and generalization. It does not give us such a connection for individual models, and we are not aware of any extant proposal that delivers this. This shift in the object of the simplicity preference is the central observation of the section, and it sets up the justificatory question we pursue in §\ref{sec:JustNewOccam}.
 
\subsection{The basic strategy}
 
The strategy consists of three steps. First, to identify as the crucial component an \emph{implicit inductive bias}  (§\ref{sssec:implicindbias}); second, to interpret this implicit inductive bias as a \emph{simplicity} bias (§\ref{sssec:implicindbiassimpl}); and finally to commit to a claim that such a simplicity bias is \emph{good} for generalization (§\ref{sssec:implicindbiassimplgood}).
 
\subsubsection{Implicit inductive bias}\label{sssec:implicindbias} 
 
\citet[p.\ 3]{ZBHRV17iclr} already write:
\begin{quote} [\dots]\ it is certainly the case that not all models that fit the training data well generalize well. Indeed, in neural networks, we almost always choose our model as the output of running stochastic gradient descent. Appealing to linear models, we analyze how SGD acts as an implicit regularizer. For linear models, SGD always converges to a solution with small norm. Hence, the algorithm itself is implicitly regularizing the solution. \end{quote}
 
In the interpolation regime, we have a textbook example of inductive underdetermination. There are several models that all fit the data perfectly but disagree on the unseen data: which one to pick? The learning algorithm must choose, and Zhang et al.\ show that for linear regression with $d \geq n$ features, SGD provably picks out the unique minimum $\ell_2$-norm interpolator among the solutions to ERM. At least for linear models, SGD uses a ``minimal-norm'' preference to break the tie.
 
\citet[p.\ 217]{Bel21ac} similarly highlights that any specific ``algorithmic'' ERM, like SGD, must go beyond Vapnik's ``algorithm-independent'' ERM paradigm and implement a way of breaking ties between minimum-training-error models---that is, in the interpolation regime, between interpolating models. He posits that in cases of benign interpolation ``an appropriate notion of functional smoothness plays a key role,'' and formulates the ``guiding principle'' (ibid., p.\ 218):
 
\begin{quote}
``\emph{Select the smoothest function, according to some notion of functional smoothness, among those that fit the data perfectly}.''
\end{quote}
 
Belkin gives the example of kernel machines, which were studied in the context of benign interpolation by \citet{BelMaMan18icml}. Kernel machines can be seen as linear regressors in higher-dimensional Hilbert spaces, and the optimization problem is to find the lowest-norm interpolator according to the norm of the space.\footnote{See, e.g., \citep[ch.\ 16]{ShaBen14} for more on kernel machines. In the interpolation regime, the optimization problem corresponds to that of the ``hard'' support vector machine paradigm (ibid., ch.\ 15).} Kernel machines can be implemented by a version of SGD, and \citet{BelMaMan18icml} suggest that understanding the inductive bias of kernel machines can shed light on the inductive bias in deep learning.
 
In the course of their analysis, \citet[p.\ 3]{BelMaMan18icml} make a useful terminological distinction, which we adopt, between \emph{regularization} and \emph{inductive bias}. The former refers to classical regularization, where fit with the training data is sacrificed for lower complexity (as in §\ref{sssec:regular}). The latter they reserve for the bias or preference required to break the tie among interpolating models.\footnote{This distinction is not quite the same as the classic one between restriction and preference biases (or \emph{soft inductive biases}, \citealp[p.\ 3]{Wil25icml}). 
Characteristic of the setting of benign interpolation is that there is apparently no restriction bias to speak of, and inductive bias as we use the term here is a type of preference bias, but so is regularization.} In the case of benign interpolation, there is apparently no regularization (no sacrifice of fit), but there must be a particular inductive bias. Further, in deep learning at least, this inductive bias is \emph{implicit}, because the learning algorithm was not explicitly designed to have it.\footnote{Note that \citet{ZBHRV17iclr} talk about ``implicit regularization,'' but what they mean, in the terms we adopt here, is implicit inductive bias. Their choice of terminology is a little confusing, because in an earlier section on ``implicit regularizations'' they do appear to discuss regularization in our sense, and dismiss both ``explicit and implicit regularizers'' as the ``fundamental reason for generalization'' (ibid., sect.\ 3.1). As mentioned in footnote \ref{fn:neysha}, earlier work \citep{NeyTomSre15iclr} hypothesized that the learning is ``implicity biasing us towards low-norm models,'' but this ``real inductive bias'' is there apparently still understood as a regularizer.} The first step of the basic strategy is thus to identify this implicit inductive bias, in particular as a preference for low-norm solutions, or for functional smoothness.

\subsubsection{Implicit inductive bias towards simplicity}\label{sssec:implicindbiassimpl} 
 
The next step interprets the inductive bias as a \emph{simplicity} bias. \citet[p.\ 18]{Bel21ac} writes that ``the idea of maximizing functional smoothness subject to interpolating the data'' is an instance of the principle that ``the simplest explanation consistent with the evidence should be preferred'' and so ``represents a very pure form of Occam's razor'' (also see \citealp[p.\ 15850]{BHMM19pnas}).
 
\citet{BarMonRak21ac} do not use the label of Occam's razor, but their analysis also leans on an inductive bias towards ``simple'' functions. As above, they observe that learning algorithms must have some bias to break the tie between interpolating solutions, and discuss examples where ``gradient methods, suitably initialized, return the empirical risk minimizers that minimize certain parameter norms'' (ibid., sect.\ 3).\footnote{They also use the terminology of ``implicit regularization,'' implicit because ``this bias is a by-product rather than an explicitly enforced property'' (ibid., p.\ 109).} They provide an in-depth analysis of benign interpolation for examples of high-dimensional linear regression (ibid., sect.\ 4) and linearized two-layer neural networks (ibid., sect.\ 6).\footnote{To motivate their focus on linear models despite the fact that deep neural nets are non-linear in their parameters, \citet[p.\ 155]{BarMonRak21ac} give two reasons. The ``direct'' reason is the existence of ``training regimes in which an overparametrized neural network is well approximated by a linear model''; the ``indirect'' reason is that ``insights and hypotheses arising from the analysis of linear models can provide useful guidance for studying more complex settings.''} Their observation is a ``simple-plus-spiky decomposition'' of the function learned in the interpolation regime. The interpolating solution is a sum of two components: one that ``is simple in a suitable sense (for instance, it is smooth)'' and one that ``is spiky: it has large complexity and allows interpolation of the data'' (ibid., p.\ 156). The simple component is ``useful for prediction'' and the spiky component is ``useful for overfitting'' (ibid., p.\ 109), where the latter ``ensures interpolation without hurting prediction accuracy'' (ibid., p.\ 128). This is again an instance of the basic strategy, postulating an implicit inductive bias toward simplicity to break the tie between interpolating solutions. This simplicity bias must not be confused with regularization in the terminology adopted above. In finding the preferred interpolating solution the algorithm does not reduce fit with training data; rather, the algorithm is set up to prefer the simplest solution among a wide range of perfectly fitting ones.
 
A different viewpoint, yet one that also conforms to this basic strategy, is offered by certain Bayesian analyses. \citet{WilIzm20nips} consider benign interpolation from the perspective of Bayesian deep learning, where a prior is formulated over the parameters of a deep neural net. They argue that ``generalization depends on \emph{two} properties: the \emph{support} and the \emph{inductive bias} of a model'' (ibid., p.\ 1). The support refers to which functions (via their parameters) 
are at all included in the prior; ``we want the support of the [model class] to be large so that we can represent any [model] we believe to be possible'' (ibid., p.\ 2). But it is also important how the prior is divided over the functions:
this is what they call the inductive bias. The combination of wide support and a ``reasonable'' inductive bias is taken to explain both the overfitting (enough support to cover even random noise) and the benign aspect (an inductive bias that among the interpolating solutions prefers good ones) of benign interpolation (ibid., sects.\ 6--7). 
 
What is a ``reasonable'' inductive bias? Wilson and Izmailov reproduce the experiments of \citet{ZBHRV17iclr} using Gaussian processes with radial basis function kernels, which ``have large support, and are thus flexible, but have inductive biases towards very simple solutions'' (ibid., pp.\ 2--3). \citet{Wil25icml} postulates more generally that
 
\begin{quote}
``in order to reproduce benign interpolation, we just need a flexible [model] space, combined with a loss function that demands we fit the data, and a simplicity bias: amongst solutions that are consistent with the data (i.e., fit the data perfectly), the simpler ones are preferred'' (p.\ 6).
\end{quote}
 
Wilson, building on work by \citet{GolFinRowWil24icml}, casts the inductive bias in neural networks as a ``bias for low Kolmogorov complexity'' \citeyearpar[p.\ 5]{Wil25icml}.\footnote{In particular, Wilson \citeyearpar[sect.\ 3.1]{Wil25icml} discusses what is essentially SRM with Kolmogorov-complexity-based weights (in particular, theorem 3.1 is the ``MDL''-SRM bound given by \citealp[sect.\ 7.3]{ShaBen14}), which invites the question whether this is not an instance of regularization.} 
\citet[p.\ 1]{MinReeValLou25nc} similarly employ a Bayesian lens to expose ``a specific Occam's razor-like inductive bias towards (Kolmogorov) simple functions'' in order to account for the generalization 
of neural nets.\footnote{Both \citep[p.\ 5]{Wil25icml} and \citep[p.\ 7]{MinReeValLou25nc} further make the connection to Solomonoff's \citeyearpar{Sol64ic} theory of universal prediction, which is based on Kolmogorov complexity and usually presented as giving some foundation for Occam's razor \citep{Ste16pos}. See \citep{Ste26arxiv} for a critical review of this theory, including of its potential for helping us understand deep learning.} Again, via the postulation of a simplicity prior, the basic strategy of these Bayesian approaches is to explain benign interpolation through an implicit inductive bias towards simpler interpolating solutions, while offering a wide support within which these simple solutions can be found.

\subsubsection{Implicit inductive bias towards simplicity is good}
\label{sssec:implicindbiassimplgood} 
 
From the story so far, it is starting to make sense that larger, more overparametrized models can be better. As \citet[p.\ 219]{Bel21ac} explains, as we move beyond the interpolation threshold in the double-descent curve, we consider larger and larger function spaces, and ``larger spaces will generally consider `better' functions.'' If $\mathcal{H}_1 \subset \mathcal{H}_2$, the smoothest or simplest interpolating solution in $\mathcal{H}_2$ will be at least as smooth or simple as the one in $\mathcal{H}_1$, and possibly even simpler.
 
This all presupposes, however, that such simpler models are indeed ``better,'' that the inductive bias towards simplicity ``is the `right' inductive bias'' \citep[p.\ 219]{Bel21ac}. That this is so, that this inductive bias is a \emph{good} bias, is the final step in the basic strategy. The full explanation offered by the strategy depends on it. 


\subsection{The new Occam's razor}\label{ssec:newoccam}

The explanation of benign interpolation arising from the basic strategy takes the following form.\footnote{One could perhaps with equal justice call this a \emph{complexity} principle, since it relies on a highly complex model class for the interpolation. We will just follow the literature here and treat it as a putative simplicity principle.}
 
\begin{quote}
\textbf{Explanation (new).} Our learning algorithm generalizes well, because it has a complex model class and selects interpolating models which are simple.
\end{quote}
 
This is meant to supersede the classical explanation from §\ref{subsec:explaining} for cases of benign interpolation. It is, like the classical one, an explanation in terms of a simplicity preference, 
a form of Occam's razor.\footnote{The 
    new explanatory principle has a direct normative counterpart: to generalize well, a learning algorithm should have a large model class and select interpolating models which are simple. That benign interpolation is often discussed in explanatory rather than normative terms reflects the fact that the inductive bias is \emph{implicit}: the learning algorithms in question were not intentionally designed to have this bias. That does not mean practitioners cannot now start to design for it. Indeed, \citet[p.\ 5]{Wil25icml} talks about ``a \emph{prescription} for building general-purpose learners.''  (Cf.\ \citealp[sect.\ 4.2]{Bucxxejps}.)} 
But this disguises a shift that creates an explanatory gap.\footnote{Some
    authors suggest a continuing role for the classical analysis in explaining benign interpolation. \citet[p.\ 91]{BarMonRak21ac} write that ``[c]lassical statistical learning theory explains the good predictive accuracy of the simple component.'' 
    More generally, \citet[p.\ 137]{HarRec22} write that ``some mathematical progress has been made to understand how deep learning leverages classical foundations of generalization.'' Even so, it seems to us that a call on the basic strategy, and a subsequent explanatory gap, remains.}

 \section{The explanatory gap} \label{sec:JustNewOccam}
 
The basic strategy of \S\ref{sec:implicit} identifies an implicit inductive bias toward simple individual models, but it does not explain why such a bias is good for generalization. In this section, we first analyze in more detail the explanatory gap that is therefore left (\S\ref{ssec:thegap}). We then sketch and comment on some general approaches towards bridging this gap (\S\ref{ssec:fillgap}). We do not endorse any particular approach: our main point is that an explanatory gap still exists.

\subsection{The gap}\label{ssec:thegap}
The gap left by the basic strategy is the final step, to explain why the simplicity bias is a good bias (\S\ref{sssec:implicindbiassimplgood}). One might wonder, though, whether we are now not merely rehearsing a familiar point about machine learning. The lesson of the ``no-free-lunch'' theorems is that every learning algorithm must possess a specific inductive bias, which prompts an explanatory or justificatory question why this inductive bias is good \citep{SteGru21syn}. What is new?

\subsubsection{The role of inductive bias}
Of course, SLT does not escape the lesson that every learning algorithm must have an inductive bias which is good in some domains but not others. In particular, as we discussed in \S\ref{ssec:srm}, hypothesis class $\mathcal{H}$ for ERM represents an inductive bias (essentially, that some $h \in \mathcal{H}$ are good), and a particular ERM$_\mathcal{H}$ will have a bad approximation error (hence true risk) in some possible domains (where no $h \in \mathcal{H}$ is good).\footnote{The no-free-lunch theorem for SLT stated by \citet[thrm.\ 5.1]{ShaBen14} says that for any learning algorithm $A$ (including ERM with particular $\mathcal{H}$) there is a distribution such that with high probability it learns a classifier with high expected error. A no-free-lunch result can also be given for regularization (ibid., remark 7.2).} 

This valid lesson might give the impression that domain-general assertions about successful learning are out of reach: good generalization just depends on the adequacy of the indispensable domain-specific inductive bias \citep{Wol96nc}. But that is too quick. As discussed by \citet{SteGru21syn},  learning theory can offer general guarantees for general algorithms which hold for a range of domain-specific inductive biases or model classes: namely, guarantees about finding the best in any given ``inductive model'' (here, model class). In particular, as we saw, SLT offers such general reliability guarantees for the general ERM and SRM algorithms.

It is true that these guarantees are only part of the story if we seek to explain ERM's or SRM's good generalization in the sense of low true risk or test error. To complete such an explanation in any particular instance, we would also need to explain why the particular inductive bias in question (choice of model class or model class sequence) is good. Nevertheless, as discussed, the theory gives us two general methodological norms, which do not rely on particular domain-specific inductive biases; and these can function as general explanations of good generalization in the sense of a small generalization gap. This is markedly different for the new explanation of \S\ref{ssec:newoccam}, which crucially relies on the goodness of a special inductive bias: an assumption that some individual models are better than others (\S\ref{sssec:implicindbiassimplgood}).

\subsubsection{A simplicity bias} \label{ssec:evensimpl}
The second step of the basic strategy was to interpret this special inductive bias as a simplicity bias: simpler models are better. In principle, we could  interpret any restriction or even preference inductive bias as some kind of simplicity bias, just because such an assumption restricts possibilities and so simplifies things. Perhaps that is overly cynical (we recognize the intuitive appeal of the notions discussed); but it highlights an important question about the work that such an interpretation actually does. There is namely also a distinct risk attached to a simplicity interpretation and the label of Occam's razor: it can suggest an explanatory or justificatory force that is not really there.

Consider again why the label was in order in the classical case. There we have a property of a model class, its capacity, and theorems connecting low capacity to generalization. Low capacity looks enough like the intuitive content of Occam's razor, fewer ways of fitting arbitrary data, that calling it a kind of simplicity is apt; and because the theorems robustly connect the very thing we are calling simple to generalization, the label licenses treating the resulting norm as a form of Occam's razor. The name reports a connection we have.

In the new case the naming runs the other way. We start from a mathematical property of individual functions, low norm or smoothness or low Kolmogorov complexity, call it simplicity, and use the name to suggest that preferring it is an application of Occam's razor. But as we discussed in \S\ref{sec:classsimplicity}, notions of the complexity of individual models are relative to a choice of language 
and there is no theorem that we know of connecting any such notion to generalization, so that 
it is not clear why \emph{these} properties should tie to extrapolating, predicting, or generalizing well.\footnote{\label{fn:specialist} A specialist may wonder whether theorems connecting individual models to generalization already exist. Norm- and margin-based bounds (e.g.\ \citealp{BarFosTel17nips}) may look like candidates, but they use the trained model's weight norm to single out a \emph{class} and prove uniform convergence over it, so the object connected to generalization is again a collection of models. Furthermore, in the interpolation regime such bounds are typically vacuous \citep{NagKol19nips}. Another candidate is the family of bounds that attach to the individual model, such as the PAC-Bayes compression bounds (\citealp{ZhoVeeAusAdaOrb19iclr,LotFinKapPotGolWil22neurips}), which are non-vacuous and are presented as a form of Occam's razor. However, these are governed by a prior and are tight only when the models that fit the data are compressible under it, so the connection to generalization runs through an assumption about the learning problem rather than through the bound alone. \citet{LotFinKapPotGolWil22neurips} supply that assumption by appeal to the ``tremendous amount of structure'' (p.~9) in real-world datasets, which is an assumption of just the sort \S\ref{sec:JustNewOccam} is about. We do not claim to have surveyed every result, but we have found none supplying, for individual models, a connection comparable to what the classical theory supplies for model classes.}

The classical theory backed two general methodological simplicity principles, independent of domain-specific inductive biases. In the new case, the Occam label might suggest the deployment of a similar methodological principle, but as a matter of fact denotes a special inductive bias, a special and substantive assumption about what models are good. The gap in the basic strategy is a story on why this assumption is justified.

\subsection{Closing the gap}\label{ssec:fillgap}
We will now sketch some ways one might go about motivating this inductive bias, going along with the idea (despite the reservations voiced in the previous section) that it is a \emph{simplicity} bias.\footnote{In particular, the language-relativity of notions of simplicity of individual models means that we do not merely need an account why (say) low-norm models are more likely to be good, but more precisely why low-norm models \emph{in a particular parametrization} are good. In other words, the choice of language is part of the inductive bias.}

\subsubsection{The flat-footed view: the world is simple}\label{ssec:ontological}
A first possibility is to see the simplicity bias as (an instance of) a fundamental principle of induction. For instance, \citet{Che25nous} proposes a  ``principle of nomic simplicity'' (PNS) to the effect that simpler propositions are more likely to be laws, and that ``should serve as a fundamental epistemic principle underlying inductive reasoning about physical reality'' (ibid., p.\ 971). Accepting that PNS cannot be ultimately justified, he suggests a transcendental argument that ``[a]t some point, we must adopt fundamental epistemic principles that explain how and why induction works,'' and ``that PNS is a good candidate for such a fundamental posit'' (ibid., p.\ 978).\footnote{Chen, who has also collaborated with Belkin \citep{CheBelBerDan26nat},  has connected this idea to the case of benign interpolation in recent presentations.} 

Chen focuses on a simplicity preference as a fundamental normative principle; a related move is to subscribe to an assumption that the world is simple (cf.\ \citealp{LinTegRol17josp}). The explanation for simpler models being better then is that the world generates data sets that exhibit the relevant kind of simplicity. This of course relies on strong metaphysical assumptions that seem hard to credit at that level of generality. While we might take the success of machine learning to constitute evidence for an empirical claim that the world is simple, this goes in the opposite direction of what we are here interested in, namely explaining this success. For this, the flat-footed claim that the world is simple does not appear to take us very far.

\subsubsection{A constructivist turn: the data are simple}
Another approach to closing the gap shifts more of the weight onto our practices. Instead of pointing to the world as the ultimate explanation for the efficacy of simplicity assumptions, we can instead say that some explanatory work is done by the process through which we produce data about the world, i.e., how we organize and classify structures in the world, collect data from them, and collate these into datasets. On this view the availability of simple patterns in the data is, at least in part, our own doing.%
    \footnote{There is a long and rich philosophical debate, going back all the way to Kantianism, over the relation between theories and data, in which the insight that scientists construct data to serve as evidence for or against theories plays a central role. See, for example: \citet{Chang2004,Steinle2016,Massimi2022}.}
Such an explanation sits well with the philosophy of science literature on data-intensive science, in which the importance of obtaining data sets that allow for uniform and standardized treatment is widely recognized \citep[e.g.][]{Leonelli2016}.

Importantly, the idea is not that in constructing the data we impose specific patterns onto the data by hand. Machine learning works in the absence of explicit feature construction. For instance, image recognition systems merely take RGB values on a grid as input, leaving the machine itself to identify the salient patterns. The data-construction view thus involves something weaker than hand-imposed patterns. We need only suppose that the process of collecting and collating data somehow results in datasets that have simple patterns embedded in it, which the machine can subsequently find by confronting the data with a very large collection of possible patterns and a bias towards simple ones. The explanatory format proposed is an interplay: a machine that can try out many patterns fast, and a dataset constructed to be amenable to this kind of pattern finding.

Note also that this view does not preclude the world from doing some of the work. For one, that same world includes the data construction processes and the scientists involved in it. Moreover, that the data harbor simple patterns may still be caused partly by structures in the world. Of some structures in the world, like natural language as it appears on the internet, we can hardly speak of data construction at all: the machines just gobble up all content that is somehow tagged as linguistic expression, so that our construction efforts only involve tagging it as such.%
    \footnote{This reminds of ``The Bitter Lesson'' \citep{Sutton2019}, a well-known criticism of anthropomorphism within AI. It is tempting to think of AI as somehow parallel to our own cognitive powers. But what explains its success might be alien to our own understandings, relying instead on brute force.}
Foreshadowing the next subsection, how large a role the data construction plays is therefore dependent on context and application domain. Nonetheless, if we bring our own efforts in collecting and curating data into view, we have more context and practice to work with when explaining the predictive success of the simplicity assumptions.

\subsubsection{Retreat: local assumptions}\label{sssec:retreat}
Until now we have mostly treated the simplicity bias as an assumption that is supposed to hold ``globally,'' across learning tasks and domains, in line with the idea that the new Occam's razor is part of a general (to repeat Belkin's phrase) ``new theory of induction.'' But this is already controversial. \citet{TenJiaGogAbb25cvpr} argue that ``[c]ontrary to popular belief, the simplicity bias [\dots]\ is not universally useful.''\footnote{Specifically, based on earlier work \citep{TenNicHarAbb24cvpr}, they argue that the simplicity bias stems from standard ReLU activation functions, and that such ReLU networks are ``near-optimal'' for image classification tasks, but not adequate for other tasks, like learning from tabular data. Also already see \citep{Sha22colt}: ``benign overfitting is implicitly `biased' towards certain learning problems, in the sense that its existence on one learning problem precludes its existence on other learning problems.''} 



This points to investigations into the shape of and motivation for the simplicity bias in restricted domains. \citet{BatWoo25arx} offer a developed instance. They argue that the generalization puzzle stems from the ``assumption (or rather a lack of assumptions) SLT makes about the data,'' and focusing on image classification, that ``real world images on which DDNs successfully generalize, conform to very specific, non-arbitrary probability distributions'' (p.\ 3). Specifically, they argue that these distributions are characterized by higher-order correlational structure and power-law scaling, and that deep networks succeed on images because they are well-suited to discovering and exploiting this structure. 

Note that such attempts to flesh out the simplicity bias as a locally valid assumption can both call upon the world (the specific domain) and our data practices (in that domain). For instance, the structure that does the explanatory work in Batterman and Woodward's account is a feature of the datasets, not of the world taken neat; but it is in the data because the world put it there: ``worldly facts are responsible for robust statistical properties that are present in the datasets upon which the DNNs are trained'' (ibid., p.\ 35).\footnote{This is worth distinguishing from a stronger appeal to our practices. On a data-construction reading, structure ends up in the data because we do a great deal of processing to put it there. On Batterman and Woodward's reading, we do comparatively little: the relevant structure is already present at the meso-scale, and the data inherit it.}
Moreover, these local assumptions could still be quite general: Batterman and Woodward leave open that the relevant ``\emph{universal} power law scaling'' (ibid., p.\ 23) is present in data sets in other domains than image classification. 


Still, such work does signal something of a retreat. Rather than upholding a simplicity principle on a par with the generality of the simplicity norms from SLT, the project becomes more of a piecemeal investigation of what inductive biases hold in what domains, whether towards simplicity, smoothness, or something else. It is indeed not clear whether Batterman and Woodward would still see the ``worldly structure'' they identify as a \emph{simplicity} bias (ibid., pp.\ 19ff).

\section{Conclusion}

Statistical learning theory gives precise conditions for successful generalization. Contemporary neural networks generalize well outside of these conditions. We have surveyed responses to this puzzle in the computer science  literature. The proposals share a common move: they locate the explanation in a preference for simpler models among the many that fit the data perfectly.
 
It is tempting to see Occam's razor at work here, but this disguises a shift. The classical theory delivers a notion of simplicity that applies to model classes, and a proof connecting that notion to generalization. The new accounts deliver a notion of simplicity that applies to individual models, but no such proof. This leaves a gap. We have surveyed some possible and ongoing moves in computer science and philosophy to close it, but as it stands the debate is far from settled.
 

\small

\bibliographystyle{abbrvnat}

\end{document}